\documentclass[runningheads]{llncs}

\usepackage[misc,geometry]{ifsym}
\usepackage{graphicx}
\usepackage{amsmath}
\usepackage{amssymb}
\usepackage{xcolor}
\usepackage{booktabs}
\usepackage{tabularx}
\usepackage{graphics}
\usepackage{hyperref}

\usepackage{cite}

\newcommand{\VSDFjaplaGZSL}{36.57}
\newcommand{\VSDFXLjaplaGZSL}{41.31}

\newcommand{\OSOCRXLjaplaGZSL}{30.83}

\newcommand{\OSOCRXLjaplaOSTR}{58.57}
\newcommand{\OSOCRXLjapreOSTR}{24.46}
\newcommand{\OSOCRXLjapprOSTR}{93.78}
\newcommand{\OSOCRXLjapfsOSTR}{38.80}

\newcommand{\ORJPfullGZSLatOverallla}{40.96}

\newcommand{\ORJPfullOSTRatOverallla}{71.86}
\newcommand{\ORJPfullOSTRatOverallre}{69.72}
\newcommand{\ORJPfullOSTRatOverallpr}{90.86}
\newcommand{\ORJPfullOSTRatOverallhm}{78.90}
\newcommand{\ORJPfullXLGZSLatOverallla}{42.58}

\newcommand{\ORJPfullXLOSTRatOverallla}{72.33}
\newcommand{\ORJPfullXLOSTRatOverallre}{72.96}
\newcommand{\ORJPfullXLOSTRatOverallpr}{92.62}
\newcommand{\ORJPfullXLOSTRatOverallhm}{81.62}

\definecolor{Cat}{HTML}{9F9BF9}
\newcommand{\cgone}[1]{\textcolor{black}{#1}}
\newcommand{\cgtwo}[1]{\textcolor{black}{#1}}
\newcommand{\cgiii}[1]{\textcolor{black}{#1}}
\newcommand{\cgiv}[1]{\textcolor{black}{#1}}
\newcommand{\cgv}[1]{\textcolor{black}{#1}}
\newcommand{\cgvi}[1]{\textcolor{black}{#1}}

\newcommand{\StSpace}[1]{\mathcal{#1}}
\newcommand{\StSet}[1]{\mathbf{#1}}
\newcommand{\Stfn}[1]{{\mathtt{#1}}}
\newcommand{\Stdom}[1]{\mathbb{#1}}
\newcommand{\Starr}[1]{\boldsymbol{#1}}
\newcommand{\Stmat}[1]{#1}

\newcommand{\Streal}[1]{#1}

\newcommand{\StGt}[1]{#1^{*}}

\newcommand{\indexof}[1]{{[#1]}}

\newcommand{\uprofit}[1]{#1_{m}}

\newcommand{\variantof}[2]{{{#1}}_{{#2}}}

\newcommand{\idxof}[2]{{#1}_\indexof{#2}}

\newcounter{relctr} %
\everydisplay\expandafter{\the\everydisplay\setcounter{relctr}{0}} %

\AtBeginDocument{}

\newcommand{\Ntesamcnt}{\Streal{N}}
\newcommand{\Nsamid}{\Streal{i}}

\newcommand{\Ntime}{t}

\newcommand{\Nsideinfo}{I}

\newcommand{\Npredtime}{\hat{\Ntime}}

\newcommand{\Nmaxtime}{\uprofit{\Ntime}}

\newcommand{\Ngttime}{\StGt{\Ntime}}

\newcommand{\Nsunk}{\variantof{s}{unk}}

\newcommand{\Nimage}{img}

\newcommand{\Ncktest}{\StSet{C}^{k}_{test}}

\newcommand{\Ncutest}{\StSet{C}^{u}_{test}}

\newcommand{\Nseq}{\Starr{Y}}

\newcommand{\Ngtseq}{\StGt{\Nseq}}

\newcommand{\Npredseq}{\hat{\Nseq}}

\newcommand{\Npredstringof}[1]{\idxof{Pred}{#1}}
\newcommand{\Ngtstringof}[1]{\idxof{Gt}{#1}}

\newcommand{\Npredat}[1]{\idxof{\Npredseq}{#1}}

\newcommand{\Nfeatdim}{\Streal{d}}

\newcommand{\Nproto}{P}

\newcommand{\Nfeatatt}{\Stmat{A}}

\newcommand{\Nfeatattat}[1]{\idxof{\Nfeatatt}{#1}}

{}

\newcommand{\Nsamtofeat}{\Stfn{B}}

\newcommand{\Nfeatseq}{F}
\newcommand{\Nfeatat}[1]{\idxof{\Nfeatseq}{#1}}

\newcommand{\Nospredictor}{\Stfn{C}}

\newcommand{\Nrec}{RE}

\newcommand{\Npre}{PR}

\newcommand{\Nfm}{FM}

\newcommand{\NLineACR}{LA}

\newcommand{\Ncharspace}{\StSpace{C}}

\newcommand{\Ntempspace}{\StSpace{T}}

\newcommand{\Rtempsof}[1]{}
\newcommand{\Ttempsof}[1]{set}

\newcommand{\Ranytemps}[1]{}
\newcommand{\Sanytemps}[1]{\Stdom{R}^{32\times 32}}
\newcommand{\Tanytemps}[1]{set}

\newcommand{\Nanychar}[1]{c}
\newcommand{\Ranychar}[1]{}
\newcommand{\Sanychar}[1]{}
\newcommand{\Tanychar}[1]{character}
\newcommand{\Danychar}[1]{A character on the $\Ncharspace$}

\newcommand{\Ranytemp}[1]{}
\newcommand{\Sanytemp}[1]{\Stdom{R}^{32\times 32}}
\newcommand{\Tanytemp}[1]{tensor}
\newcommand{\Danytemp}[1]{A template on the template space $\Ntempspace$. }

\newcommand{\Ranyproto}[1]{}
\newcommand{\Sanyproto}[1]{\Stdom{R}^\Nfeatdim}
\newcommand{\Tanyproto}[1]{tensor}
\newcommand{\Danyproto}[1]{}

\newcommand{\Rcharfeatmap}[1]{}
\newcommand{\Scharfeatmap}[1]{\Stdom{R}^{w_g \times h_g \times \Nfeatdim^{'}}}
\newcommand{\Tcharfeatmap}[1]{tensor}
\newcommand{\Dcharfeatmap}[1]{An (intermediate) feature map of the input glyph.}

\newcommand{\Rglyphfeatmap}[1]{}
\newcommand{\Sglyphfeatmap}[1]{\Stdom{R}^{w_g \times h_g \times \Nfeatdim^{'}}}
\newcommand{\Tglyphfeatmap}[1]{tensor}
\newcommand{\Dglyphfeatmap}[1]{An (intermediate) feature map of the input glyph.}

\newcommand{\Nfeatmap}[1]{\variantof{M}{{#1}}}

\newcommand{\Rfeatmap}[1]{}
\newcommand{\Sfeatmap}[1]{\Stdom{R}^{w \times h \times \Nfeatdim^{'}}}
\newcommand{\Tfeatmap}[1]{tensor}
\newcommand{\Dfeatmap}[1]{An (intermediate) feature map of the input word clip.}

\newcommand{\Nsamefn}{\Stfn{Same}}

\newcommand{\Nfnrej}{\Stfn{Rej}}

\newcommand{\Nctxat}[1]{c}
\newcommand{\Nctxestat}[1]{\hat{c}}

\newcommand{\Ndispatcher}{\Stfn{D}}
\newcommand{\Nexpertset}{\StSet{E}}
\newcommand{\Nactiveexperts}{\variantof{\Nexpertset}{a}}
\newcommand{\Nexpert}{\Stfn{E}}
\newcommand{\Nexpertof}[1]{\variantof{\Nexpert}{#1}}
\newcommand{\NfeatseqMOE}{\StSet{F}}
\newcommand{\Naspectratio}{r}
\newcommand{\Naspectrange}[1]{(\variantof{r^{#1}}{lwr},\variantof{r^{#1}}{upr}]}
\newcommand{\Nexptemplate}[1]{\variantof{T}{#1}}
\newcommand{\Naggregator}{\Stfn{A}}

\newcommand{\Ncrossent}[2]{\Stfn{CrossEntropy}({#1},{#2})}

\newcommand{\Tname}{Open-Set Text Recognition}

\newcommand{\texttempinf}{{temporal information}}

\newcommand{\textMOSE}{MOoSE}
\newcommand{\textMoSe}{Multi-Orientation Sharing Experts}

\newcommand{\textMoOsTr}{Multi-Oriented Open-Set Text Recognition}
\newcommand{\textMOOSTR}{MOOSTR}%

\newcommand{\mkRQ}[2]{{\color{#2}\textbf{RQ{#1}}}}

\newcommand{\RQone}{\mkRQ{1}{black}}
\newcommand{\RQtwo}{\mkRQ{2}{black}}
\newcommand{\RQthree}{\mkRQ{3}{black}}
\newcommand{\RQfour}{\mkRQ{4}{black}}

\newcommand{\mooseRQone}{\textit{Vertical samples are only a small percentage of the dataset. Would ignoring them be a simpler approach with minimal effect on total accuracy?}}

\newcommand{\mooseRQtwo}{\textit{Can we simply rotate the vertical samples line like~\cite{fudan23_moocr}?}}

\newcommand{\mooseRQthree}{\textit{Is using only one single horizontal expert like~\cite{sandem} sufficient?}}

\newcommand{\mooseRQfour}{\textit{What components of the model should we share?}}

\newcommand{\RQoner}{\par \mkRQ{1}{black}: \mooseRQone{} \newline}
\newcommand{\RQtwor}{\par \mkRQ{2}{black} \mooseRQtwo{} \newline}
\newcommand{\RQthreer}{\par \mkRQ{3}{black} \mooseRQthree{} \newline}
\newcommand{\RQfourr}{\par \mkRQ{4}{black} \mooseRQfour{} \newline}

\newcommand{\LongRQ}[1]{
	\vspace{-0.8em}
	\begin{itemize}
		\item[*] {#1}
	\end{itemize}
	\vspace{-0.8em}
}

\newcommand{\RQonel}{\LongRQ{ \mkRQ{1}{black}: \mooseRQone{}}}
\newcommand{\RQtwol}{\LongRQ{ \mkRQ{2}{black} \mooseRQtwo{} \newline}}
\newcommand{\RQthreel}{\LongRQ{ \mkRQ{3}{black} \mooseRQthree{}}}
\newcommand{\RQfourl}{\LongRQ {\mkRQ{4}{black} \mooseRQfour{}}}

\newcommand{\etal}{\textit{et al.}}

\begin{document}

\title {\textMOSE{}:  \textMoSe{} for Open-set Scene Text Recognition}
\titlerunning{Accepted in ICDAR 2024}

\authorrunning{C. Liu, S. Corbillé, and EH. Barney Smith}

\author{Chang Liu \and Simon Corbillé  \and Elisa H Barney Smith}
\institute{Machine Learning Group, 
	\cgvi{Luleå University of Technology}, Sweden \\
\email{chang.liu@ltu.se, simon.corbille@associated.ltu.se, \newline elisa.barney@ltu.se }}
\maketitle

\begin{abstract}
	Open-set text recognition, which aims to address both novel characters and previously seen ones, is one of the rising subtopics in the text recognition field.
	However, the current open-set text recognition \cgv{solutions} only focuses on horizontal text, which fail to model the real-life challenges posed by the \cgone{variety of} writing directions \cgone{in} real-world scene text.
	Multi-orientation text recognition, in general, faces challenges from the diverse image aspect ratios, significant imbalance in data amount, and domain gaps between orientations.  %
	In this work, we first propose a \textMoOsTr{} task (\textMOOSTR{}) to model the challenges \cgv{of} both novel characters and writing \cgone{direction} variety. 
	We then propose a \textMoSe{} (\textMOSE{}) framework as a strong baseline \cgv{solution}. \textMOSE{} uses a mixture-of-experts scheme to alleviate the domain gaps between orientations, while \cgone{exploiting} common \cgone{structural} knowledge among experts to alleviate the data scarcity that some experts \cgone{face}.
	The proposed \textMOSE{} framework is validated by ablative experiments, and also tested for feasibility on the existing open-set benchmark. Code, models, and documents \cgv{are} available at: \url{https://github.com/lancercat/Moose/}
	
\end{abstract}


\section{Introduction}
\begin{figure}[t]
	\centering
	\includegraphics[width=\linewidth]{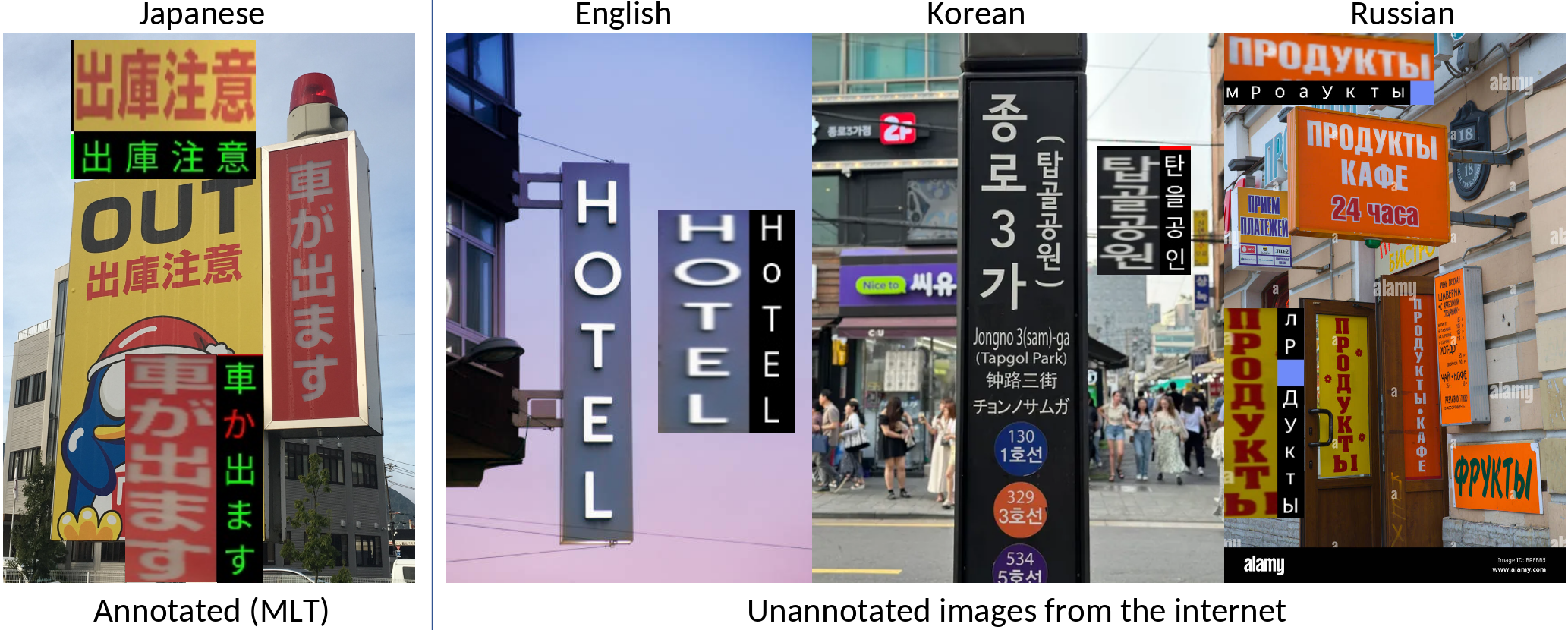}
	\caption{Samples of multi-orientated text instances in the wild \cgv{from various sources}, showcasing the multi-orientation recognition capability of the proposed \textMoSe{} framework. The model is \textbf{only} trained with English and Chinese data.
 	\cgv{The Japanese sample comes from the annotated MLT dataset \cgvi{and} recognition, and thus results are shown in colors. Green and red characters indicating correctly and wrongly predicted characters. The rest \cgvi{of the} images \cgvi{come} from the internet, so the results are shown in white, and a blue block indicates rejection. }}
	\label{fig:comp}
\end{figure}

The open-set text recognition task,  as an emerging subfield of text recognition, aims to address the presence of novel characters and contextual biases \cgone{introduced by \cgv{the use of multiple language scripts} and language evolution~\cite{imtok}}. 
Specifically, for novel characters the task first requires the model to spot samples that include ``out-of-set" characters for human inspection. Then the model needs to incrementally acquire the  capability \cgvi{to recognize} these characters without retraining, by registering the side-information~\cite{neko20nocr,neko23osavr}.
Open-set text recognition is well-suited to scene text recognition scenerios, where characters from a new script can be frequently encountered as in the Korean text image in Fig.~\ref{fig:comp} which includes text in 4 languages (Korean, English, Chinese, and Japanese) in one single scene.

\cgv{In additional to multiple scripts, the images in Fig.~\ref{fig:comp} contain text from multiple orientations,} like many close-set scene text recognition benchmarks~\cite{iiit5k,cute,svt}, the open-set text recognition task~\cite{neko20nocr} fails to model the orientation variety of the real-life multilingual data.  
Inspired by Yu \etal{}'s close-set multi-orientation benchmark~\cite{fudan23_moocr}, we propose a \textMoOsTr{} task (\textMOOSTR), to model \cgv{the} challenges from \cgv{both} orientation variety and novel \cgv{characters} in the real-world scene text images.

To address the vast differences in sample aspect ratio when vertical and horizontal text are both considered, several methods rotate vertical samples and treat them as horizontal text~\cite{stride,fudan23_moocr}, yielding orientation-agnostic character representations. 
\cgv{Most existing methods either resize variouse images to a same-size~\cite{aon}, or rotate vertical images~\cite{fudan23_moocr}, which would be either \cgvi{computationally} inefficient or lead to missing orientation information.} Inspired by the anchor mechanism~\cite{faster}, we propose a \textMoSe{} framework~(\textMOSE{}), providing a unified solution to \cgone{the challenges of both orientation and aspect-ratio diversity}, by adopting a Mixture-of-Expert~\cite{imoe} scheme. 

Implementation-wise, we split \cgone{text samples into three groups, \textit{i.e.},} long horizontal, short horizontal, and vertical text.  \cgone{Each group is resized and padded to fit a dedicated template, and routed to its dedicated expert.} The design allows us to preserve the original orientation \cgone{of a sample}, while avoiding incurring large computing overheads caused by excessive scaling or padding. In addition, we propose a partial sharing scheme among different anchors to \cgone{exploit} common structural knowledge between orientations~\cite{titiccv21}, while staying robust against domain gaps.

\cgiii{Experiment-wise, we justify our framework by answering the following research questions. First we validate multi-orientation text recognition itself, \textit{i.e}:}

\RQonel{}
\cgiii{We then validate the feasibility \cgv{of} handling multi-orientation text under \cgv{the} open-set context, by comparing it against several existing methods that are \cgv{used} on 
\cgv{the} close-set \cgv{problem}~\cite{sandem,fudan23_moocr}, specifically:}
\RQtwol{} \vspace{-4mm}\RQthreel{}
 We also run extra examination on sharing policies among the experts, discussing which module should be shared across experts or not, \textit{i.e.}: 
 \RQfourl{}
The contributions of this work can be summarized as the following:
\begin{itemize}
	\item \cgiv{We propose} a \textMoOsTr{} task which expands existing tasks \cgone{to cover both challenges from orientation variety and novel characters in real world scene text data.}
	\item  \cgiii{We propose} a \textMoSe{} framework that is capable of handling seen and unseen text written in various orientation forming a baseline of the task.
	\item \cgiii{We propose} a scheme that allows us to divide samples in a more fine-grained fashion, exploiting common structural knowledge between orientations.
\end{itemize}

\section{Related Work}
\subsection{Open-Set Text Recognition}
The open-set text recognition~(OSTR) task~\cite{neko20nocr} models the evolving character set and context in a dynamic environment.
OSTR aims to find ``out-of-set'' characters that fail to match any registered centers~\cite{icdar23osr}, and adapt the model(s) to recognize the new characters incrementally without retraining, by mapping the side-information describing the character to build new class centers~\cite{jinic21,cm19,eccv20}.

The OSTR task is naturally suited to multi-script scene text recognition tasks, because it allows characters from new scripts to be actively spotted and gradually added as they are encountered. 
However,  current open-set text recognition task setups and methods~\cite{neko20nocr,vsdf,neko23osavr} focus only on the classification behavior, and completely ignore the variety of orientations in real-life scene text data, hence limiting their feasibility. In this work, we propose to expand the task into a more realistic multi-orientatation setup by introducing vertical text samples and providing a strong baseline method suited to them.

\begin{figure}[h]
	\centering
	\includegraphics[width=0.7\linewidth]{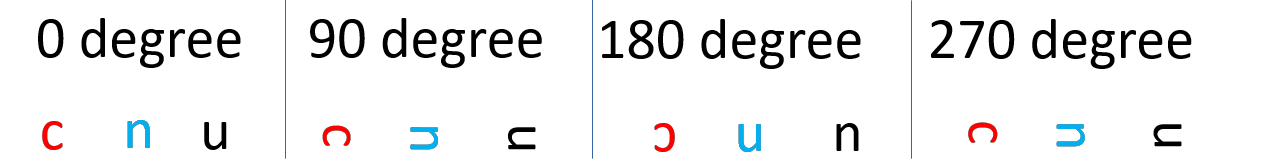}
	\caption{Examples of characters that are ambiguous when orientation information is not available \cgiii{(`c', `n', and `u' in this case)}.}
	\label{fig:confuse}
\end{figure}

\subsection{Multi-Orientation Text Recognition}
Though irregular-shaped text recognition is a popular research topic~\cite{trocr,moran,ASTER}, multi-orientation text recognition, as an important subfield of scene text recognition,  is long overlooked due to the scarcity of vertical Latin text samples~\cite{fudan23_moocr}.
Current works on multi-orientation text recognition can be generally categorized according to whether or not the orientation of the input images are preserved.

Specifically, one category~\cite{stride,accessdvi,fullshare} prefers to rotate the input image so that the vertical samples fit into the horizontal batch, which may end up learning orientation invariant character representations. However, in open-set text recognition tasks where characters are distinguished by their visual shape~\cite{vsdf}, this trait could be infeasible. In addition to `c',`u', and`n' already illustrated in Fig.~\ref{fig:confuse}, other characters such as `$\mathcal{E}$' and `m', `{\fontfamily{qcr}\selectfont
	I}' and `H' are hard to distinguish without knowing the orientation out-of-context.

The second category of approaches to multi-orientation text recognition preserves the input orientation as is. A trivial way to do this is to scale all images to a square~\cite{aon}, however, this yields extra overhead to the model as the square needs to be large enough to fit the content from samples from diverse aspect ratios. 
To alleviate this issue, methods like SAN~\cite{sandem} choose to resize the vertical samples and horizontal samples into different sizes for cotraining. However, the vast varying sample aspect ratios within the horizontal samples causes these methods also to face excessive computational costs from over-stretching like AON~\cite{aon} and can limit their performance.
To further reduce computation cost overhead introduced by stretching and padding, this proposed work further splits the horizontal data into short samples and long samples.

\section{\textMoOsTr{}}
In this work, we propose a \textMoOsTr{} (\textMOOSTR) Benchmark to cover both of the variety of writing order challenge and the novel character challenge that are common in real-world scene text data. The task is an expansion of the Open-set text recognition task~\cite{neko20nocr}, \cgv{to include} vertical samples.

\begin{figure*}[t]
	\centering
	\includegraphics[width=\linewidth]{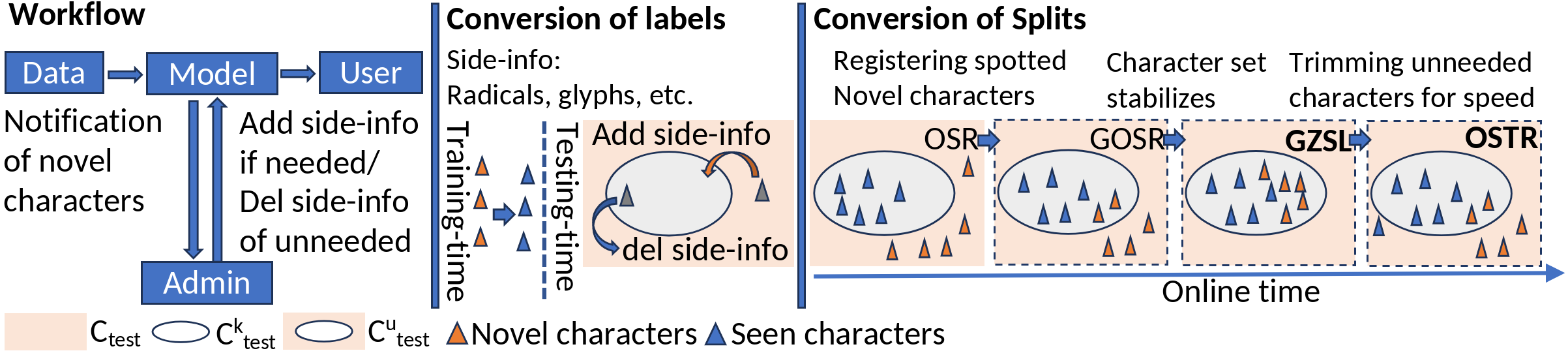}
	\caption{The work flow, label conversions, splits conversions of the \textMOOSTR{} task.}
	\label{fig:osocr101}
\end{figure*}

\textMOOSTR{} has four splits: \cgv{Genrealized} Zero-shot Learning~\cite{gzsl-survey} (GZSL), \cgv{Open-Set Recognition}~\cite{tosr} (OSR), Generalized Open-Set Recognition~\cite{osr-survey} (GOSR), and Open-Set Text Recognition~\cite{neko20nocr} (OSTR) modeling different stages from  open-environment use cases, illustrated in Fig.~\ref{fig:osocr101}.
After training the model is configured with a set of seen characters $\Ncktest$ and deployed to a volatile data stream, corresponding to the OSR split. 
When \cgone{an image} sample containing novel characters appears, the model is supposed to inform the administrator to inspect the sample (by rejection). 
If the admin decides that the novel characters are of interest, the model is expected to be adaptable to recognize the character by registering its side information without retraining, corresponding to the GOSR split. 

As the \cgone{character-set growth goes on}, the character set may stabilize for a certain period, corresponding to the GZSL split, which is similar to other zero-shot text recognition tasks like~\cite{jinic21,zhang20pr,eccv20}. \cgone{Note for some use cases, e. g., data involving emojis, the character-set may not stabilize, getting stuck in the GOSR stage.}

Alongside the evolution, some characters that come from the initial configuration or that are added halfway through can appear very rarely.
In speed critical cases, the model is expected to support removing the registered character to speed up the process, and raise a notification if the removed characters appear again, \cgone{corresponding to the OSTR split}.

In summary, the use case can be described with two core \cgvi{functionalities}: matching and rejection. Similar to the OSTR task, the \textMOOSTR{} task demands the model recognize a sample with all its characters in $\Ncktest$, i.e., with associated side-information registered to the model, and reject samples with one or more characters that are not in $\Ncktest$.

Four metrics are used to evaluate the algorithm performance. The recognition performance is measured by Line Accuracy (LA),
\begin{equation}
	\NLineACR{}=\frac{1}{\Ntesamcnt}\sum_{\Nsamid=1}^{\Ntesamcnt} \Nsamefn(\Ngtstringof{\Nsamid},\Npredstringof{\Nsamid}).
\end{equation}
The rejection performance is measured by Recall and Precision,
\begin{equation}
	\begin{split}
		\Nrec=&\frac{\sum_{\Nsamid=1}^\Ntesamcnt{} \Nfnrej(\Npredstringof{\Nsamid})\Nfnrej(\Ngtstringof{\Nsamid})}{\sum_{\Nsamid=1}^N \Nfnrej(\Ngtstringof{\Nsamid})}, \Npre=\frac{\sum_{\Nsamid=1}^\Ntesamcnt{} \Nfnrej(\Npredstringof{\Nsamid})\Nfnrej(\Ngtstringof{\Nsamid})}{\sum_{\Nsamid=1}^N \Nfnrej(\Npredstringof{\Nsamid})},\\
	\end{split}
\end{equation}
and are summarized with the F-score\footnote{\cgvi{In this paper, the styles of math symbols indicate their types: $\Stfn{X}$ indicates functions, $\StSet{X}$ indicates sets, $X$ indicates variables, and $\Starr{X}$ indicates arrays, $x$ indicates single numbers.} },
\begin{equation}
	\Nfm=\frac{2\Nrec*\Npre}{\Nrec+\Npre}.
\end{equation}

\begin{table*}[t]
	\centering
	\caption{Dataset scales and character-set sizes. Italic indicates novel characters.}
	\begin{tabular}{l|p{1.2cm}p{1.2cm}||p{1.2cm}p{1.2cm}|p{1.2cm}p{1cm}p{1cm}|p{1cm}}
		\hline
		Name&Training samples & Num. classes & Testing set & Num. classes &  Shared Kanji & \textit{Unique 
		Kanji} &  Latin &  \textit{Kana}  \\ \hline
		OSTR~\cite{neko20nocr}&~~379K &3791& 4009&1435 &~~787 &~452 & 36 & 160   \\ 
		\textMOOSTR{}&~~468K &3791& 5074 &1543 &~~849 &~497 & 36 & 161 \\ 
		\hline
	\end{tabular}
	\label{tab:taskcomp}
	\vspace{-1.3em}
\end{table*}

Data-wise, the training samples include English and Chinese word crops taken from ART~\cite{art}, RCTW~\cite{rctw}, CTW~\cite{ctw}, LSVT~\cite{lsvt}, and the Latin and Chinese section of the MLT~\cite{mlt19} \cgone{dataset}. Note that the samples that \cgone{include} characters other than the 3755 Tier1 Chinese characters, 26 English letters, and the 10 digits \cgv{(3791 classes in total)} are excluded from the training set to avoid label leaking.
The testing set includes the Japanese subsets of the MLT dataset. 
A comparison between the two benchmarking datasets is shown in Table~\ref{tab:taskcomp}.

\section{Method}

\begin{figure}[b]
	\centering
	\includegraphics[width=\linewidth]{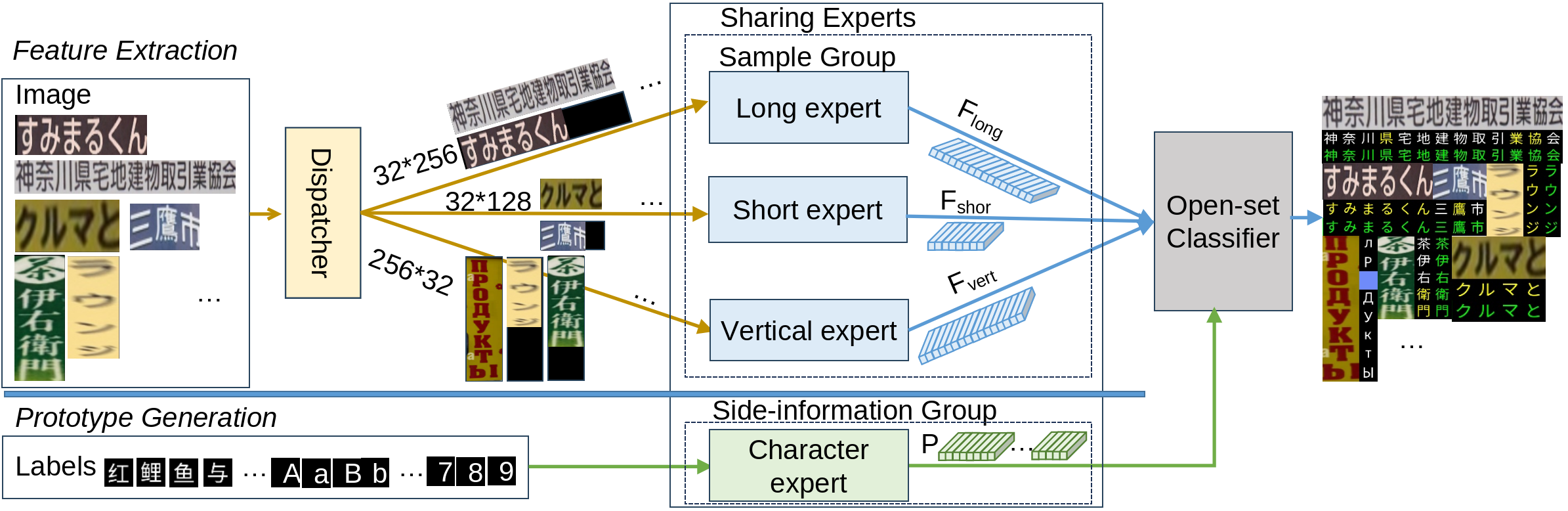}
	\caption{The proposed \textMoSe{}~(\textMOSE) framework. Images are dispatched to the appropriate expert based on their orientation determined from the aspect ratio. \cgv{In additional, a character expert is used to generalize class-centers for the open-set classifier.}}
	\label{fig:framework}
	\vspace{-1.3em}
\end{figure}

Like objects, text in the real world has a vast diversity of aspect ratios~\cite{ham} (see Fig.~\ref{fig:comp}). In this work, we propose a \textMoSe{} framework (\cgone{\textMOSE{}}), shown in Fig.~\ref{fig:framework} to address the challenge caused by different text orientations in the context of open-set text recognition.

The \textMOSE{} framework is composed of three parts: the dispatcher, the \cgv{Sharing E}xperts, and the open-set classification module, which are described in detail in the following subsections. 

\subsection{Dispatcher}
The dispatcher $\Ndispatcher$ takes each input image $\Nimage$ and decides the subset of experts $\Nactiveexperts \subset \Nexpertset$ which should be utilized to handle the image: 
\begin{equation}
	\Nactiveexperts=\Ndispatcher{}(\Nimage), \NfeatseqMOE =\cup_{\Nexpertof{i}\in \Nactiveexperts} \Nexpertof{i}(\Nimage) .\\	
\end{equation}

\begin{figure}[t]
	\centering
	\includegraphics[width=\linewidth]{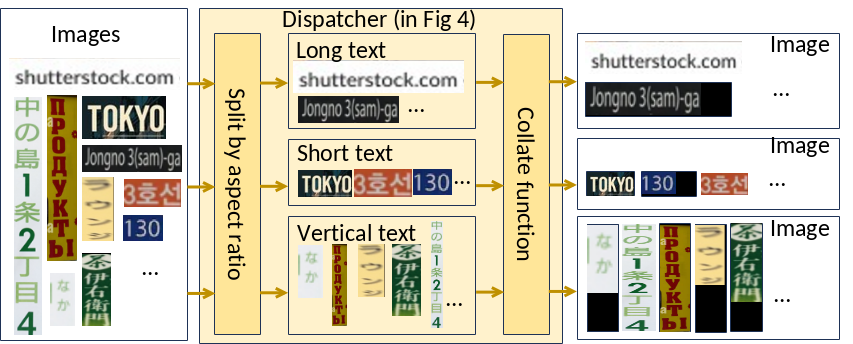}
	\caption{The proposed sample dispatcher \cgv{(The yellow block in Fig~\ref{fig:framework})}. \cgone{The module routes the input images to different experts for feature extraction according to the aspect ratio of the image.} }
	\label{fig:dispatcher}
\end{figure}

In this work the dispatcher (Fig.~\ref{fig:dispatcher}) is implemented as a rule-based module that sends text crops to corresponding experts according to their aspect ratio.
We first designate each expert $\Nexpert$ with a non-overlapping aspect ratio interval $\Naspectrange{\Nexpert}$ and \cgone{a size} template $\Nexptemplate{}$. The dispatcher first selects active experts with regards to the aspect ratio $\Naspectratio:=\frac{w}{h}$ of the input image:
\begin{equation}
	\Nactiveexperts=\left\{\Nexpert \in \Nexpertset | \Naspectratio \in \Naspectrange{\Nexpert} \right\} .
\end{equation}
After selecting the active experts, the dispatcher resizes and pads the image according to the associated templates for feature extraction. The collate function resizes the image \cgone{while keeping} its aspect ratio, \cgone{such that} the short side \cgone{of the resized image matches the short edge of the corresponding size template. If the longer edge of the resized image is shorter than the size template, it gets padded to the template size. Otherwise, it's streched to fit the template.}

\begin{figure}[b]
	\centering
	\includegraphics[width=\linewidth]{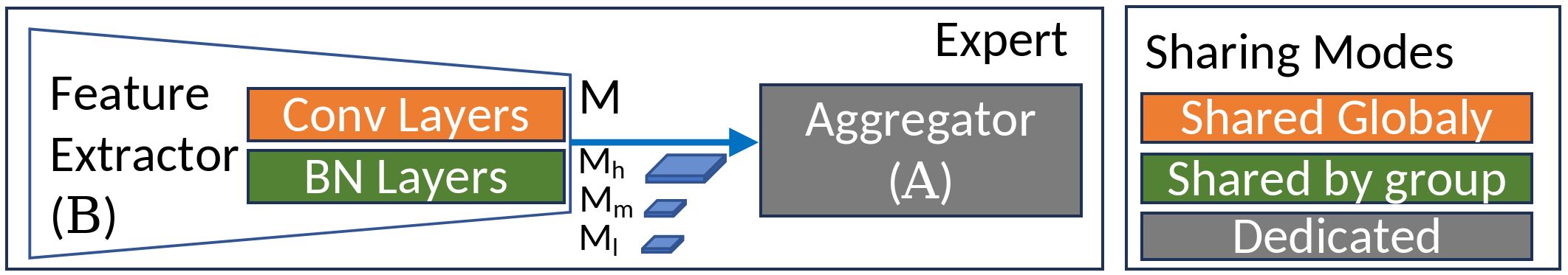}
	\caption{An expert is composed of a feature extractor and an aggregator, where the convolution layers in the feature extractors are shared among experts, and the batch norm layers remain domain dedicated.  }
	\label{fig:expert}

\end{figure}

\subsection{Sharing Experts}
\cgiii{An expert converts the input images into features on the feature space, specifically, an expert $\Nexpertof{}$ is \cgv{a} differentiable function} that extract character features from an input image:
\begin{equation}
	\Nfeatseq=\Nexpert{}(\Nimage).
\end{equation} 
The experts are implemented as a combination of a shared feature extractor $\Nsamtofeat$ and an aggregator $\Naggregator$ (Fig.~\ref{fig:expert}).
The feature extractor \cgone{encodes the input image into a feature pyramid} $\Nfeatmap{}:(\Nfeatmap{l}, \Nfeatmap{m}, \Nfeatmap{h})$:
\begin{equation}
	\Nfeatmap{}=\Stfn{\Nsamtofeat}(\Nimage),
\end{equation}
\cgone{where} $\Nsamtofeat$ is implemented as a DSBN-ResNet45~\cite{neko23osavr} \cgone{Network. In this work}, all word-level experts share the same set of \cgv{batch normalization} layers, while the character expert for prototype \cgone{generation} has its own set of \cgone{dedicated} batch normalization layers following~\cite{vsdf}.

The aggregator $\Naggregator$ is then used to sample the feature \cgone{maps from}~$\Nfeatmap{}$ into character features $\Nfeatseq$:
\begin{equation}
	\Nfeatseq=\Naggregator(\Nfeatmap{}) .
\end{equation}

\cgiii{In this work, experts are grouped into Sample Group and Side-information Group by functionality. Experts in the Sample Group extracts time-stamp aligned feature sequence from input word images, while Side-information group generates normalized prototypes (class-centers) w.r.t. the given side-information~\cite{neko20nocr}.}

\begin{figure}[b]
	\centering
	\includegraphics[width=\linewidth]{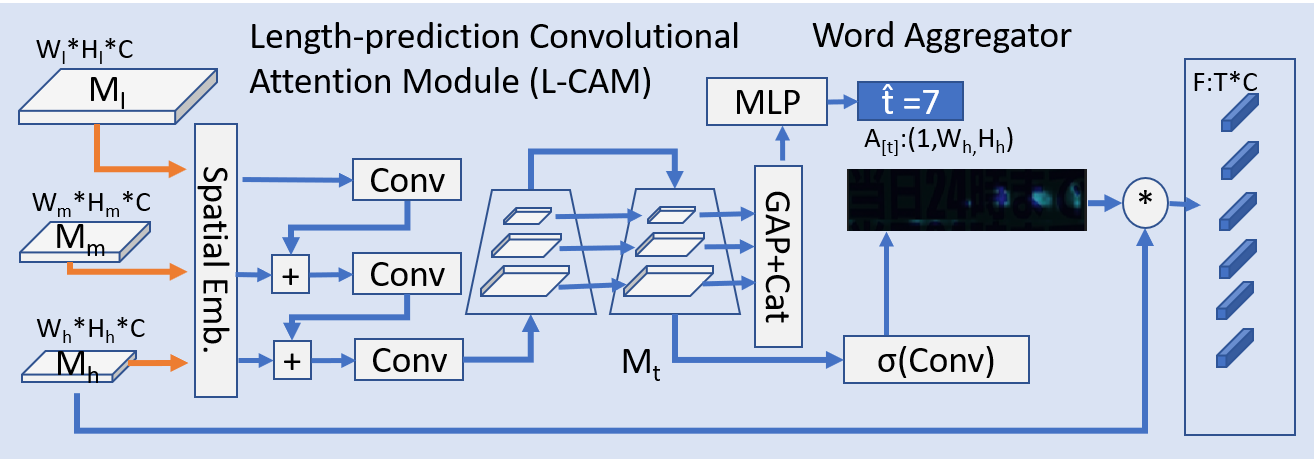}
	\caption{\cgv{The aggregator (see Fig.~\ref{fig:expert}) implementation for sequence data. This implementation is adopted by experts in the Sample Group (\cgvi{blue ones} shown in Fig~\ref{fig:framework}). }}
	\label{fig:lcam}
\end{figure}

\cgvi{\subsubsection{The Sample Group}includes experts that extract time-stamp aligned character features from the input images. In this work, we adopt three word-image associated experts, namely the Long Expert, Short Expert, and Vertical Expert (Blue blocks in Fig.~\ref{fig:framework}). Each expert adopts a L-CAM module~\cite{vsdf} with extra spatial embedding~\cite{dino} as the aggregator (Fig.~\ref{fig:lcam}). }

\cgvi{Specifically, the module \cgone{first captures} global \texttempinf{} of the input feature \cgone{maps} via an FPN structure.}
\cgone{The module then aggregates the temporal information with a global average pooling layer (GAP),} and predicts the sequence length $\Npredtime$ via an MLP layer. \cgone{The feature map extracted with FPN is also used to generate location masks of each possible timestamp $\Nfeatatt{}: (\Nfeatattat{1},...,\Nfeatattat{\Nmaxtime})$  with a convolution layer}. 

\begin{figure}[b]
	\centering
	\includegraphics[width=0.7\linewidth]{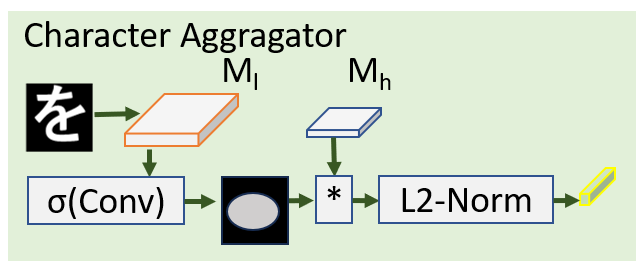}
	\caption{\cgv{The aggregator (see Fig.~\ref{fig:expert}) implementation for individual character data. This implementation is adopted by experts in the Side-information Group (\cgvi{the green one} shown in Fig~\ref{fig:framework}).}}
	\label{fig:caggr}
\end{figure}

\cgone{Finally, features of characters in the image $\Nfeatseq$ are} sampled from the high-level feature map $\Nfeatmap{h}$ with attention operation:
\begin{equation}
	\Nfeatat{t}=\sum_{i,j=0,0}^{i,j=w,h} \Nfeatattat{t,i,j}\Nfeatmap{h} .
\end{equation}
During training, $\Nfeatseq$ is trimmed with the ground truth of the sequence length. \cgone{During evaluation, it is truncated according to the predicted length $\Nmaxtime{}$.}
\cgiii{\subsubsection{The Side-information Group} \cgvi{includes} a single character expert $\variantof{\Nexpert}{c}$} \cgvi{(Green block in Fig.~\ref{fig:framework})}, which implements the capability to incrementally recognize novel characters~\cite{neko20nocr,neko23osavr}. Specifically, the character expert dynamically \cgvi{builds} class centers (prototypes) $\Nproto$ from the provided side-information $\Nsideinfo$:
\begin{equation}
	\Nproto=\variantof{\Nexpert}{c}(\Nsideinfo).
\end{equation}
For this expert, a character aggregator (Fig.~\ref{fig:caggr}) is used to sample the \cgone{character for feature maps extracted from its glyph image. The module first predicts the foreground mask} of the character w.r.t the low-level feature map $\Nfeatmap{l}$ \cgone{with a sigmoid-convolution block. It then aggregates the high-level feature map \cgv{$\Nfeatmap{h}$} into a raw character representation, using the predicted mask as attention weights. Finally, the raw representations are normalized by length \cgiii{(i.e. $|\Nproto|$)}, resulting in the generated prototypes.} 

\cgiii{Following~\cite{vsdf,neko23osavr}, the character expert has its dedicated \cgv{batch normalization} weights. The sharing policy among the Sample Group (\RQfour) is discussed and validated in Sec.~\ref{sec:ablsharing} in \cgv{detail}.}

\subsection{Open-Set Classification}
To implement the capability to reject out-of-set characters~\cite{neko20nocr,neko23osavr}, the framework utilizes an open-set classification $\Nospredictor$, which matches the prototypes $\Nproto$ and the extracted character features $\NfeatseqMOE$ to produce prediction result $\Npredseq$:
\begin{equation}
	\Npredseq=\Nospredictor(\NfeatseqMOE,\Nproto).
\end{equation}
\cgone{Since} in this work each image will only be assigned to one expert, there will always be one \cgone{single} element in $\NfeatseqMOE$, so there is no need to aggregate the output of multiple experts.  

\cgiii{Hence, we simply adopt the open-set classifier in~\cite{neko23osavr} as \cgvi{the} classification function $\Nospredictor$, specifically:
\begin{equation}
	\begin{split}
		\Npredat{\Ntime}:=&\Stfn{Softmax}(\left[\Nfeatat{\Ntime}\Nproto^{T},\Nsunk|\Nfeatat{\Ntime}|\right]),
	\end{split}
\end{equation}
where $\Nfeatat{\Ntime}$ is the feature extracted by the selected expert at the $\Ntime$-th timestamp, $\Nsunk$ is a learnable parameter as the cosine similarity threshold for rejection, and $[]$ is the Concatenation operator.}

\subsection{Optimization and Inference}

The model is optimized by minimizing the summation of the classification loss and length prediction loss from each expert,
\begin{equation}
    \begin{split}
        L=&\sum_{a\in \Nexpertset} \Ncrossent{\variantof{\Npredseq}{a}}{\variantof{\Ngtseq}{a}} +\sum_{a\in \Nexpertset} \Ncrossent{\variantof{\Npredtime}{a}}{\variantof{\Ngttime}{a}}.\\
    \end{split}
\end{equation}

Before training, each expert is connected with its dedicated datastream, generated by the dispatcher as a preprocessing step. 
The experts are then trained in a ``Soft parameter sharing''~\cite{mtlsurvey} fashion. In each training iteration, a local label set is sampled~\cite{neko20nocr} according to the ground truths in the data batch for each expert, whose corresponding glyphs are used to generate expert-specific prototypes.

\cgone{During evaluation, the character expert is no longer updated, hence the prototypes $\Nproto$ can be cached instead of dynamically generated for each sample repetitively. Prototypes for novel characters can be incrementally generated and appended by calling the character expert with their corresponding glyphs.}

\section{Experiments}

In this work three sets of experiments are conducted. We first conduct an ablative study on the GZSL split of \cgv{the} proposed \textMoOsTr{} (\textMOOSTR{}) benchmark to validate the \textMoSe{}~\cgv{(\textMOSE) framework} \cgone{design} and implementation. 
Then, we benchmark \cgone{\textMOSE{} on all four splits of the \textMOOSTR{} for feasibility.} Furthermore, \cgtwo{as a referenced comparison} we compare the proposed method with SOTAs on  \cgone{the} OSTR task~\cite{neko20nocr,neko23osavr}.

\subsection{Implementation Details}
The framework proposed in this work is implemented with Pytorch. The models are trained with an Adadelta optimizer with a learning rate of $1.0$, decaying to $0.1$ after 600k iterations. The model is trained with a weight decay of $10^{-5}$. In this section, we report the average accuracy from 100k to 400k iterations to reduce the factor of coincidency.

We \cgone{adopt} 3 word-level experts for word samples, specifically the long \cgone{expert} $\Nexpertof{long}$, the short \cgone{expert} $\Nexpertof{short}$, and the vertical expert $\Nexpertof{vert}$. The aspect ratio intervals are set as $[6,\inf)$, $[0.5,6)$, and $(-\inf,0.5)$, \cgone{and}
templates $\Nexptemplate{long}, \Nexptemplate{short}, \Nexptemplate{vert}$ are set as $32 \times 256, 32 \times 128, 128 \times 256$ respectively. 

During training the batch size of the short expert is set to 48 following OSOCR~\cite{neko20nocr}, while the batch sizes of the long and the vertical experts are set to 24, \cgone{due to their larger size templates. We let experts with} larger templates have smaller batch sizes to keep the computation costs in check.

\subsection{Ablative Experiments}
\begin{table}[t]
	\centering
	\caption{Studies on the effect of introducing vertical samples}
	\begin{tabular}{l|l|l|l|l|l}
		\hline
		Models & Vertical Handling & Experts & \textbf{LA (all)} & LA (vert) & LA(Hori) \\ \hline
		Horizontal & Dropped & Horizontal & 32.93 & 16.40 & 37.33 \\
		Single-Horizontal & Keep as is   & Horizontal, Vertical & 34.55 & \textbf{29.45} & 35.90 \\ 
		\hline
		Horizontal-MoE & Dropped & Short, Long & 33.66 & 16.82 & \textbf{38.15} \\ 
		\textbf{MOoSE} & Keep as is & Short, Long, Vertical & \textbf{35.92} & 29.18 & 37.71 \\ 
		\hline
	\end{tabular}
	\label{tab:motrade}
\end{table}
\cgiii{First, to answer \RQone{}}, we validate the trade-off between orientations of the proposed \textMOSE{} framework in Table~\ref{tab:motrade}. The results suggest that introducing vertical images for training generally hampers horizontal performance, however will overall benefit the performance \cgone{on the entire dataset}. In addition, we found that \cgv{h}aving more than one horizontal expert reduces the performance drop that \cgv{is} caused by introducing the vertical samples.

\begin{table}[b]
	\centering
	\caption{Ablative studies to evaluate anchor designs}
	\begin{tabular}{l|l|l|l}
		\hline
		Models &~Vertical Handling &~Experts &~\textbf{LA(all)} \\ \hline
		
		Horizontal-MoE &~Dropped &~Short, Long & ~33.66 \\ \hline
		Rotated &~Rotate to Horizontal~ & ~Short, Long & ~34.80 \\ 
		Single-Horizontal &~Keep as is   & ~Horizontal, Vertical & ~34.55 \\ 
		\hline
        \textbf{\textMOSE{}} &~Keep as is & ~Short, Long, Vertical~ & \textbf{~35.92} \\ \hline
	\end{tabular}
	\label{tab:abl-designs}

\end{table}

We \cgiii{then} compare \textMOSE{} with other possible designs to demonstrate the feasibility \cgv{of its anchor design and sharing policy}.  We first discuss the handling of the vertical samples  to answer \RQtwo{} \cgv{and} \RQthree. Quantitative results are shown in \cgone{Table}~\ref{tab:abl-designs}. The comparison with method ``Horizontal-MoE'' indicates that vertical samples take a significant portion of scene text, and should not be ignored \cgone{during training}. Comparison \cgone{with} the ``Rotated'' approach indicates that it is better to keep the vertical images as is, than \cgone{to rotate} the vertical samples and \cgone{treat} them like horizontal ones~\cite{fudan23_moocr}. The necessity of having two anchors for horizontal experts is validated by comparing with \cgone{the} ``Single-Horizontal'' \cgone{model,} which has a \cgv{design} similar to~\cite{sandem}.

\label{sec:ablsharing}

\begin{table}[t]
	\caption{
		Ablative results on the sharing policies}
	\begin{center}
		\begin{tabular}{l|c|c|l|l|l}
			\hline
			
			Name &Attention sharing&Backbone Sharing &\textbf{LA (all) $\uparrow$}& LA (vert) $\uparrow$& LA (hori) $\uparrow$\\
			\hline
			Share None&&&~25.45 &~18.17 &~26.10\\
			\hline
			Share All&\checkmark&\checkmark&~28.41  & ~15.31& ~31.90  \\
			\hline
			\textbf{\textMOSE}&&\checkmark&\textbf{~35.92} & \textbf{~29.18 }& \textbf{~37.71}\\
			\hline
		\end{tabular}
		\label{tab:abl-moostr}
	\end{center}
\end{table}

\cgtwo{We then compare \textMOSE{} to various sharing policies on the Multi-orientation open-set text recognition task to answer \RQfour{}, and the results are shown in Table~\ref{tab:abl-moostr}. Results indicate that sharing a convolutional backbone is better than having independent experts, or sharing everything among experts. Note that there is a significant gap between the ``Share None'' approach and the``horizontal-MoE'' approach in Table~\ref{tab:motrade} on horizontal samples. The gap is caused by the difference of whether or not the character expert shares its backbone with the experts. }

\subsection{\textMoOsTr{}}

\begin{table*}[b]
	\centering
	\caption{The performance on the \textMOOSTR{} benchmark.}
	\begin{tabular}{c|p{3cm}|p{2cm}|l|l|l|l|l}
		\hline
		Split  & Registered & Out-of-Set & Method & \textbf{LA (all)} & R & P & \textbf{FM} \\ \hline
		GZSL & Latin, Shared Kanji & ~ & MOoSE & 34.27 &~-- &~-- &~-- \\ 
		~ & Unique Kanji, Kana & ~ & \textMOSE{}-XL & \textbf{37.39} & ~-- & ~-- & ~-- \\ \hline
		OSR & Latin, Shared Kanji &  Unique Kanji& MOoSE & 69.08 &71.56 & 94.05 & 81.28 \\ 
		~ & ~ & Kana& \textMOSE{}-XL & \textbf{75.56} & \textbf{74.05} & \textbf{95.79} & \textbf{83.53} \\ \hline
		GOSR & Latin, Shared Kanji & Kana & MOoSE & 56.73 & 62.27 & 76.98 & 68.85 \\ 
		~ & Unique Kanji &   & \textMOSE{}-XL & \textbf{59.81} & \textbf{63.83} & \textbf{81.89} & \textbf{71.74} \\ \hline
		OSTR & Shared Kanji & Latin & MOoSE & 59.96 & \textbf{80.42} & 84.91 & 82.61 \\ 
		~ & Unique Kanji & Kana & \textMOSE{}-XL & \textbf{61.75} & 80.01 & \textbf{88.18} & \textbf{83.90} \\ \hline
	\end{tabular}
	\label{tab:perfmoostr}
\end{table*}

\cgone{In this part we benchmark the proposed \textMOSE{} framework to show its performance capability in the multi-orientation scenario.}  The models are trained for 200k iterations \cgone{following~\cite{neko20nocr}}, and quantitative results are shown in Table~\ref{tab:perfmoostr}. \cgiii{Besides the base model, we provide an alternative \cgv{large} model called \textMOSE-XL, which simply doubles the backbone feature channels in \textMOSE. }

Recognition-wise, results on the GZSL split indicate that the \textMOSE{} framework demonstrates some extent of generalization capability to unseen characters \cgone{in} both horizontal and vertical samples. The huge line accuracy gap between GZSL and the rest of the splits indicates that the model still struggles to generalize to \cgone{kanas (hirakatas and kataganas),} which are characters that have a strong style gap with characters from the training set.  On the other hand, the accuracies  achieved in the GOSR and OSTR split around $60\%$ indicate that the model still \cgv{generalizes} decently on novel Japanese Kanjis that have close structures to the characters in the training set. 
The OSR split indicates that the model demonstrates a decent recognition accuracy of $75.56\%$ on samples with characters seen on the training set.

\begin{figure}[t]
	\centering
	\includegraphics[width=\linewidth]{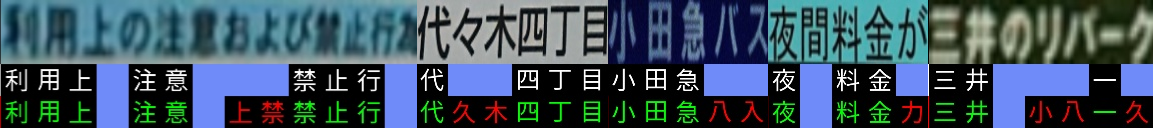}
	\caption{Failed rejection on samples with Hirakatas and Kataganas from the OSR split. White indicates seen characters, \cgv{blue} block indicates the character is novel and thus subject to rejection. Green and red indicates correct and wrong predictions.}
	\label{fig:rejection-issue-kanas}
\end{figure}

Rejection-wise, the model keeps a recall over $60\%$ and precision over $80\%$, which means it can reliably spot out-of-set characters from the data stream with a high accuracy. The recall and precision gap between the OSR and GOSR splits show the model is less capable of finding novel samples with kanas than novel kanjis.

\begin{figure}[!b]
	\centering
	\includegraphics[width=\linewidth]{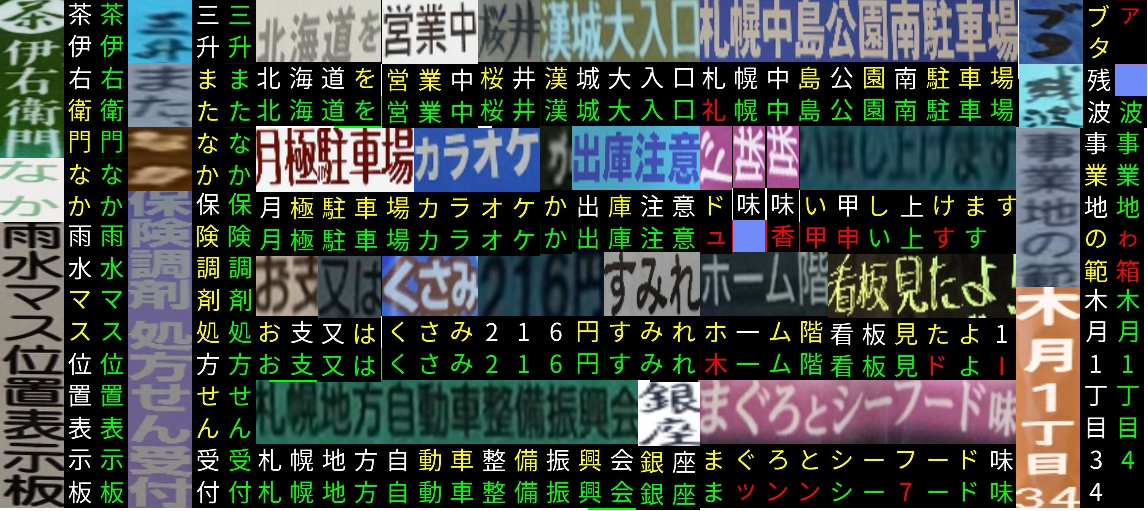}
	\caption{Qualitative recognition results from \textMOOSTR{} benchmark. Recognition results are shown with yellow label characters indicating novel characters, white characters indicating seen characters, green characters indicating correctly predicted characters and red indicating a recognition error.}
	\label{fig:perfmoostr}

\end{figure}

We further find the kanas mainly fall into the decision boundaries of seen kanjis (see Fig.~\ref{fig:rejection-issue-kanas}). Considering novel kanjis have higher recall than kanas, \cgone{which means the novel kanjis are less likely to fall in a known decision boundary. Hence} it is likely \cgv{that} the trained feature extractor \cgv{is} less generalized to kanas\cgv{, causing them to spread} wider in the feature space and hence \cgv{they} are more likely to fall in the wrong decision boundaries instead of getting rejected. 
The Recall gap between OSTR and GOSR indicates that the model could reliably reject seen characters if their side information is unregistered from the model.  The users can safely remove training-set characters if they do not frequently appear in the data, and expect a rejection to happen once they re-appear.

Qualitative-wise, we visualize the results from the GZSL split in Fig.~\ref{fig:perfmoostr}.
Results indicate decent robustness against text of various styles and orientations. However, the \cgone{aggregation model $\Naggregator$} still fails to locate some characters, especially in the later part of a long script. \cgone{The model} also shows a tendency to confuse characters with close shapes, due to insufficient modeling of detailed information of characters.

\begin{table*}[t]
	\caption{
		Performance on the open~--set text recognition task. Bold indicates the main metrics, and italic indicates characters unseen during training. Asterisk mean vertical data is also used in training.}
	\begin{center}
		\resizebox{\linewidth}{!}{
			\begin{tabular}{l|p{2cm}|p{2cm}||l|l|l|l|l}
				\hline
				Split&$\Ncktest$ & $\Ncutest$&Name&\textbf{LA}&R&P&\textbf{FM}\\
				\hline
				&\textit{Unique Kanji}&&OSOCR-Large~\cite{neko20nocr}&\OSOCRXLjaplaGZSL{}  &~--&~--&~--\\
				{GZSL}&Shared Kanji,&$\emptyset$&OpenCCD~\cite{vsdf}&\VSDFjaplaGZSL{}&~--&~--&~--\\
				&\textit{Kana}, Latin,&&OpenCCD-Large~\cite{vsdf}&\VSDFXLjaplaGZSL{}&~--&~--&~--\\
				& & &OpenSAVR~\cite{neko23osavr}&\ORJPfullGZSLatOverallla{}&~--&~--&~--\\
				&&&OpenSAVR-XL~\cite{neko23osavr}&\textbf{\ORJPfullXLGZSLatOverallla{}}&~--&~--&~--\\
				&&&MOoSE-XL*&39.56&~--&~--&~--\\
				\hline
				OSTR&Shared Kanji, &\textit{Kana}&OSOCR-Large~\cite{neko20nocr} &\OSOCRXLjaplaOSTR&\OSOCRXLjapreOSTR&\textbf{\OSOCRXLjapprOSTR}&\OSOCRXLjapfsOSTR\\
				&\textit{Unique Kanji}&{Latin} &OpenSAVR~\cite{neko23osavr}&\ORJPfullOSTRatOverallla{}&\ORJPfullOSTRatOverallre{}&\ORJPfullOSTRatOverallpr{}&\ORJPfullOSTRatOverallhm{}\\	
				&  &&OpenSAVR-XL~\cite{neko23osavr}&\ORJPfullXLOSTRatOverallla{}&\textbf{\ORJPfullXLOSTRatOverallre{}}&\ORJPfullXLOSTRatOverallpr{}&\ORJPfullXLOSTRatOverallhm{}\\	
				&&&MOoSE-XL*&64.80&\textbf{80.49}&89.12&\textbf{84.59}\\
				\hline
		\end{tabular}}
		\label{tab:perfostr}
	\end{center}

\end{table*}

\subsection{\Tname{}}
To make a referenced comparison to previous works, we also evaluate the \textMOSE{} framework with the OSTR testing protocols. Since the testing set only involves horizontal data, the vertical expert is not used in this experiment. 
Specifically, we report the model at 200k iterations following~\cite{neko20nocr,vsdf,neko23osavr}. \cgone{Results} are shown in Table~\ref{tab:perfostr}, \cgone{indicating} that the recognition performance is 3 percent lower than the SOTA approach~\cite{neko23osavr}, likely due to the lack of context separation~\cite{vsdf} and detail modeling~\cite{neko23osavr}. However, the lack of horizontal text performance is somewhat compensated for by its capability of recognizing samples from different orientations, which is an important part of real-world scene text. Hence, we still consider the proposed \textMOSE{}
as a feasible baseline model of the \textMOOSTR{} task.

\section{Limitations}
Even though the proposed \textMoSe{} framework demonstrates decent multi-orientation recognition performance, the model suffers some lower performance on the horizontal benchmarks due to lack of context separation~\cite{vsdf} and detail modeling~\cite{neko23osavr}, which will be gradually resolved in future work. 

\cgiii{Another limitation of this work is the somewhat naive design of the experts and the routing policy, which is knowledge driven and not optimal. In future works, \cgvi{we} plan to \cgv{develop} a data-driven routing approach, and also make the anchor design \cgiv{self-adaptive to the data~\cite{aan}.} }

Task-wise, the training set and the testing set mainly focus on the CJK family in the scene text context, hence lacking diversity. We are working to annotate and release a more diverse and inclusive benchmark protocol in the near future.

\section{Conclusion}
In this work, we proposed a \textMoOsTr{} benchmark~(\textMOOSTR), which takes both novel characters and diverse orientations into account. To \cgiii{implement the multi-orient text recognition capability,} we introduced a \textMoSe{}~(\textMOSE{}) framework which forms a solid baseline for this challenge. 

\cgiii{For each research questions we proposed at the beginning of this paper, we summarized their answers here: }
\begin{itemize}
	\item \RQoner{}
	\textbf{No.} While they are small, they are still a large enough portion of the corpora to require special attention. Specifically, ignoring them causes a performance drop of $ 2.26\% $ when we take this approach shown in Table~\ref{tab:abl-designs} (Row 1 vs. Row~2).
	
	\item \RQtwor{} 
	\textbf{No.} Rotating it causes the performance to decrease by $1.12$ percent, shown in Table~\ref{tab:abl-designs} (Row 1 vs. Row 3).
	
	\item \RQthreer{} 
	\textbf{No.} As shown in Table~\ref{tab:abl-designs}, having two anchors improves the average line accuracy by $1.37 \%$ percent (Row 1 vs. Row 4).

	\item \RQfourr{}
	\textbf{Just the backbone.} We explored sharing the convolutional layers and the attention module. It benefits the process if we share the convolutional layers, and keep expert-dedicated attention modules (Table~\ref{tab:abl-moostr}).

\end{itemize}

\textMOSE{} also demonstrated decent feasibility in existing open-set benchmarks~\cite{neko20nocr} and generalization capability to a wide range of \cgvi{scripts} (see Fig.~\ref{fig:comp}). Results in this paper provide a strong foundation for multi-orientation and open set recognition, on which future efforts can be built.

\section{Acknowledgement}
This work was partially supported by the Wallenberg AI, Autonomous Systems and Software Program (WASP) and the Kempestiftelserna grant CSMK23-0109. The computations were enabled by the Berzelius resource at the National Supercomputer Center.

\bibliographystyle{unsrt}  %

\bibliography{osocrsdk/citations_mk4/openocr.bib,osocrsdk/citations_mk4/closeocr.bib,osocrsdk/citations_mk4/open_this_and_that.bib,osocrsdk/citations_mk4/part_comp.bib,osocrsdk/citations_mk4/fewandzero.bib,osocrsdk/citations_mk4/wut.bib,osocrsdk/citations_mk4/this_and_that.bib,osocrsdk/citations_mk4/oovocr.bib,osocrsdk/citations_mk4/ocrdatasets.bib,osocrsdk/citations_mk4/dguda.bib,osocrsdk/citations_mk4/ilflclssl.bib,osocrsdk/citations_mk4/chs_shit.bib,osocrsdk/citations_mk4/background.bib,osocrsdk/citations_mk4/yc_5311.bib,osocrsdk/citations_mk4/discuss_1124.bib}

\end{document}